\documentclass[10pt,conference]{IEEEtran}
\IEEEoverridecommandlockouts
\usepackage{cite}
\usepackage{url}
\usepackage{amsmath,amssymb,amsfonts}
\usepackage{algorithmic}
\usepackage{algorithm}
\usepackage{graphicx}
\usepackage{multirow}
\usepackage{textcomp}
\usepackage{xcolor}
\usepackage{array}
\usepackage{threeparttable}
\def\BibTeX{{\rm B\kern-.05em{\sc i\kern-.025em b}\kern-.08em
    T\kern-.1667em\lower.7ex\hbox{E}\kern-.125emX}}
\begin{document}

\title{Lightweight, Rubric-Guided Trajectory Evaluation for Production AI Agents}


\author{\IEEEauthorblockN{Linh-An Phan\IEEEauthorrefmark{1},
MingXue Wang\IEEEauthorrefmark{1},
Guangyu Wu\IEEEauthorrefmark{1}, 
Feng Pan\IEEEauthorrefmark{2},
Zhaoyu Pang\IEEEauthorrefmark{2},
Yanbin Zhang\IEEEauthorrefmark{2}
\IEEEauthorblockA{\IEEEauthorrefmark{1}Huawei Ireland Research Center, Dublin, Ireland, \IEEEauthorrefmark{2}Huawei Shenzhen R\&D Center, China}
\{linhanphan, wangmingxue, guangyuwu, panfeng, pangzhaoyu, zhangyanbin\}@huawei.com
}
}


\maketitle

\begin{abstract}
Trajectory evaluation is essential for improving the reliability of LLM-based agents, but production use makes it expensive to run repeatedly. Modern agents generate long traces containing tool calls, observations, retries, and external outputs, while not all raw tokens are equally useful for diagnosis. We present \textit{LiteTrajEval}, a lightweight architecture for budget-bounded trajectory evaluation. LiteTrajEval derives compact domain-specific rule profiles offline, then preprocesses each trajectory online, marks heuristic failure signals, serializes it under a fixed global budget, and invokes a single rubric-guided LLM judge to produce structured diagnostic reports. Evaluated on public Magentic-One-style and $\tau$-bench-style trajectory datasets, LiteTrajEval improves failure-localization alignment with human annotations by roughly 20--35 percentage points on Magentic-One and up to 23 percentage points on $\tau$-retail compared with AgentRx, while reducing cost by about 6$\times$ and evaluation time by more than 8$\times$. This solution has also been deployed in our enterprise agentic platform.
\end{abstract}


\section{Introduction}

LLM-based agents increasingly perform complex workflows involving planning, tool use, retrieval, code execution, state updates, and multi-step decision making. For such systems, the final answer is only a partial indicator of reliability. A correct answer may hide unsafe tool use, policy violations, unnecessary retries, or poorly grounded intermediate reasoning. Meanwhile, an incorrect answer may be caused by an early misread observation, a failed tool call, missing evidence, or poor recovery after a previous error. The execution trajectory therefore becomes a central artifact for understanding agent behavior and improving agent reliability.

Unlike traditional software logs, however, agent trajectories often contain error messages that are not direct indicators of the true root cause. Agents may recover through retries or alternative tool paths, and early reasoning mistakes may only surface later as downstream tool failures, making purely rule-based diagnosis insufficient. In production agent systems, where traces accumulate continuously and evaluation must run repeatedly across prompt versions, model upgrades, and regression suites, this diagnostic challenge is compounded by a cost and latency constraint that one-off benchmark analysis does not face.

Recent work has moved from final-answer evaluation toward process-level and trajectory-level evaluation. $\tau$-bench~\cite{yao2024taubench} shows that agents can fail to follow domain policies and update external state correctly across realistic multi-turn interactions. AgentBoard~\cite{chang2024agentboard} and T-Eval~\cite{chen2024teval} evaluate agents using progress, grounding, and step-level skill signals rather than a single success metric. More directly, trajectory-diagnosis methods such as TRACE~\cite{kim2025trace}, AgentRewardBench~\cite{lu2025agentrewardbench}, Agent-as-a-Judge~\cite{zhuge2025agentjudge}, AgentDiagnose~\cite{ou2025agentdiagnose}, and AgentRx~\cite{barke2026agentrx} demonstrate that analyzing the full task-solving process can reveal failure modes that final-answer grading cannot capture. However, these methods address trajectory quality in isolation and do not account for the cost and latency demands of repeated production evaluation.

In practice, evaluation must run across prompt versions, tool implementations, model upgrades, and failure-analysis batches, so cost and token usage need to be predictable. Long-running agents may generate many reasoning steps, tool calls, observations, retries, logs, stack traces, and large external outputs. Feeding the entire raw trace into a judge is inefficient and can suffer from attention degradation in long contexts~\cite{liu2023lost}, yet aggressively truncating it can break the causal chain needed for diagnosis. Redundant tool loops, repeated observations, and verbose logs may consume most of the token budget while contributing little diagnostic signal. The central question is therefore whether a single-call, fixed-budget evaluator can preserve enough of the execution to support reliable failure localization.

\begin{table}[t]
\centering
\small
\caption{LLM-call patterns in trajectory evaluation methods.}
\label{tab}
\begin{tabular}{|l|l|}
\hline
\textbf{Solution} & \textbf{LLM calls} \\ \hline
AgentRx~\cite{barke2026agentrx} & Multiple \\ \hline
Holistic Evaluation~\cite{holistic2026} & Multiple \\ \hline
AgentEval~\cite{agenteval2026} & Multiple \\ \hline
TRACE~\cite{kim2025trace} & Multiple \\ \hline
\textbf{LiteTrajEval (our)} & \textbf{Single} \\
\hline
\end{tabular}

\vspace{0.3em}
\footnotesize{Multiple-call methods evaluate individual steps before judging trajectory quality; LiteTrajEval uses one judge call per trajectory.}
\end{table}

Many recent systems address this with multi-call LLM pipelines that evaluate individual steps or spans before aggregating a trajectory-level judgment, as shown in Table~\ref{tab}. This can be effective but increases cost and latency, making it difficult to attach to production regression workflows where thousands of trajectories may need to be evaluated.

In this paper, we present \textit{LiteTrajEval}, a lightweight architecture for budget-bounded trajectory evaluation. It separates reusable offline domain profiling from online per-trajectory diagnosis. At evaluation time, LiteTrajEval normalizes heterogeneous traces, applies heuristic rules to mark failure-oriented signals, removes or compacts redundant low-value content, preserves the global execution skeleton, and serializes the trajectory under a fixed global budget. A single rubric-guided LLM judge then produces a structured diagnostic report with candidate failure steps, root-cause explanations, failure categories, key observations, and rubric scores.

This paper makes two main contributions:

\begin{enumerate}

\item \textbf{A fixed-budget, single-call architecture for production trajectory evaluation.}
We propose LiteTrajEval, an offline-online pipeline that preserves a holistic trajectory view, bounds per-trajectory judge input, and uses one rubric-guided LLM call to produce structured diagnostics. It shows that deterministic preprocessing, status-aware compaction, and budget-aware serialization can make single-call trajectory evaluation viable without a multi-stage diagnosis pipeline.

\item \textbf{An empirical comparison against a strong failure-diagnosis baseline.} On Magentic-One-style and $\tau$-retail trajectories, LiteTrajEval improves group-level failure-localization alignment with human annotations by roughly 20--35 percentage points on Magentic-One and by up to 23 percentage points on $\tau$-retail compared with AgentRx, while reducing total evaluation cost by about 6$\times$ under the same judge model, by about 22$\times$ when using DeepSeek-v4-flash, and reduces full evaluation time by more than 8$\times$.

\end{enumerate}

LiteTrajEval is not intended to replace task-specific success metrics, constraint-based checks, or human review. Instead, it provides a deployable middle layer between final-answer evaluation and heavyweight multi-call root-cause analysis, and has been integrated into the evaluation harness of our production agentic platform.

\begin{figure*}[!t]
\centering
\includegraphics[width=0.95\linewidth]{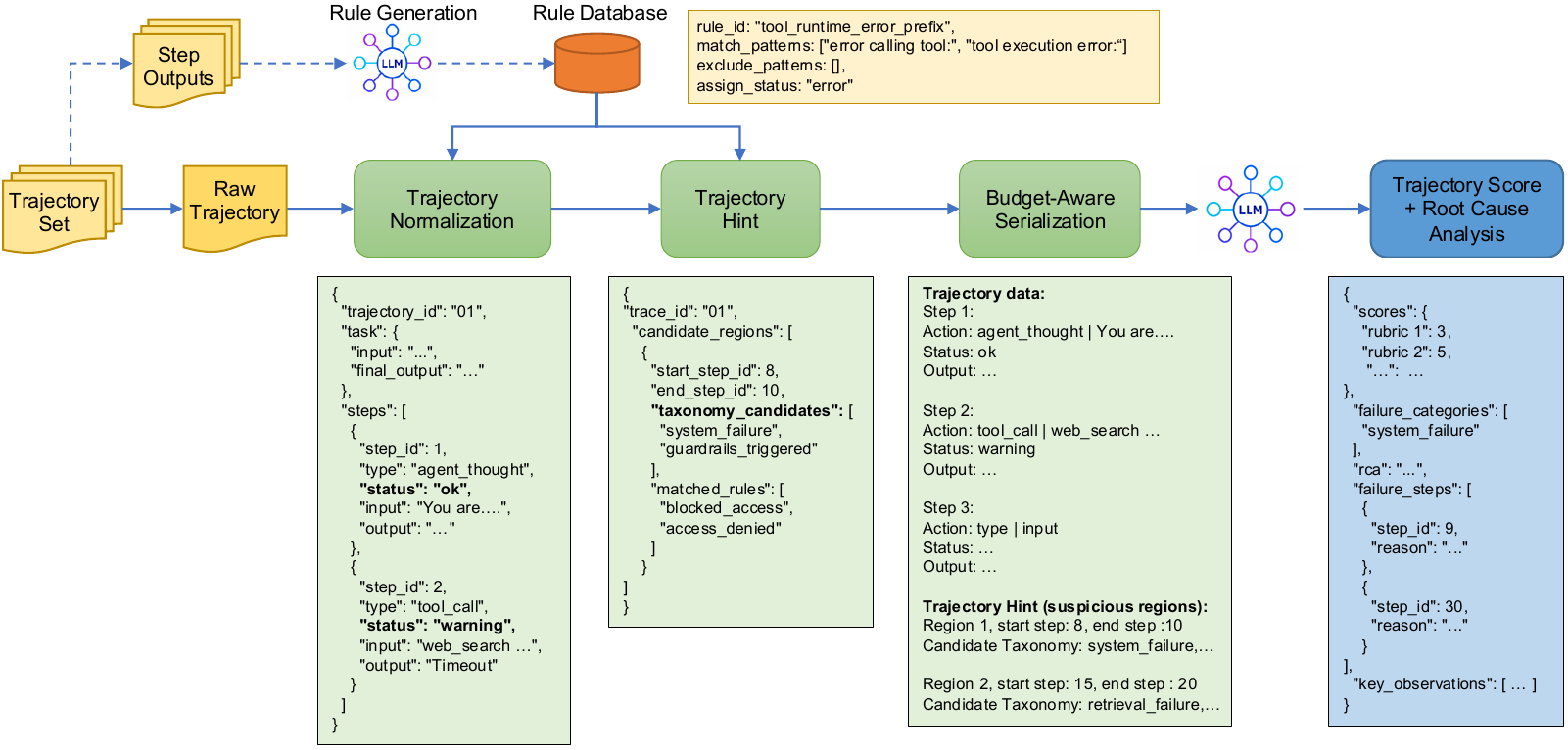}
\caption{Overview of the LiteTrajEval pipeline. An offline step (dash line) generates domain-specific rule profiles from accumulated trajectory outputs. The online path processes heterogeneous agent execution logs, serializes trajectories under a fixed budget, and evaluates them with rubric-guided LLM judge.}
\label{fig7}
\end{figure*}

\section{System Design}

LiteTrajEval is a lightweight trajectory evaluation layer for production agent workflows. It separates offline rule profiling, run once per domain, from online per-trajectory evaluation. The offline path generates domain-specific rule profiles from accumulated trajectories; the online path applies these rules, serializes each trajectory under a fixed budget, and invokes one rubric-guided LLM judge. This uses LLMs for semantic rule discovery without adding LLM calls to online preprocessing.

\subsection{Offline Rule Generation}

Rule generation operates on step outputs rather than full trajectories. LiteTrajEval uses accumulated trajectory data to produce a compact, domain-specific rule profile offline, then applies the resulting rules deterministically during online preprocessing. This offline process has four steps:

\begin{itemize}
\item \textbf{Output extraction:} extract the \texttt{output} field from each collected trajectory step, since status rules are intended to detect suspicious output patterns rather than reason over full trajectories.
\item \textbf{Text canonicalization:} normalize common volatile tokens, such as URLs, UUIDs, timestamps, file paths, hexadecimal strings, request identifiers, and long numeric values, so recurring operational patterns become easier to detect.
\item \textbf{Representative selection:} apply near-duplicate detection using Jaccard similarity over normalized token sets. If an output exceeds a threshold (e.g., 0.7), it is grouped with that representative and only its support count is updated.
\item \textbf{Rule synthesis:} chunk representative groups under a token budget and pass them with the current rule profile to an LLM that proposes or refines structured rules. This offline pass typically needs only a few LLM calls per domain; rules remain stable unless the step-output distribution changes, and can be reviewed or updated by developers.
\end{itemize}

Each generated rule contains a rule identifier, priority, scope, match patterns, optional exclusion patterns, an assigned status, and candidate taxonomy labels. Rules only assign \texttt{warning} or \texttt{error}; outputs matching no rule default to \texttt{ok}. Thus, the offline LLM call provides domain-specific semantic rule discovery, while the online path remains deterministic and efficient.

\subsection{Trajectory Preprocessing}

Given a new trajectory, LiteTrajEval first converts heterogeneous execution logs into a minimal normalized schema. Agent runtimes differ in how they expose reasoning steps, tool calls, observations, retries, errors, and final responses. The normalized schema keeps only the information needed for evaluation: the original task, the final output, and an ordered sequence of steps. Each step contains a \texttt{step\_id}, \texttt{type}, \texttt{status}, \texttt{input}, and \texttt{output}. The \texttt{status} field is inferred from the centralized rule profile. Rules detect \texttt{warning} and \texttt{error} conditions from step outputs; unmatched steps default to \texttt{ok}. This keeps online inference simple while retaining domain-specific failure signals.

After status inference, LiteTrajEval extracts lightweight failure-oriented hints. It selects warning and error steps as anchors, expands them into small local regions using nearby step order, shared tool or action type, repeated inputs, or similar input/output fingerprints, and attaches candidate taxonomy labels from the matched rules. These hints are weak guidance rather than final decisions: they help the judge attend to suspicious regions in long trajectories and make outputs easier to aggregate across traces.

The output of preprocessing is a normalized trajectory plus a compact guidance artifact containing candidate failure regions, matched rule identifiers, and candidate taxonomy labels. This stage provides a stable structure for budget-aware serialization and single-call judging.

\subsection{Evidence-Aware Budget Serialization}

Even after preprocessing, trajectories may still be too long to send directly to an LLM judge. Raw outputs can contain verbose webpage observations, large JSON payloads, stack traces, repeated tool responses, or redundant orchestration messages. LiteTrajEval addresses this by modeling trajectory serialization as a constrained optimization problem. The goal is to produce a bounded representation under a fixed global character budget $B$ while preserving the task definition, step order, and failure-relevant evidence. The online serializer is entirely deterministic.

Given a normalized trajectory \(T\), LiteTrajEval applies three deterministic refinement steps before final rendering.

\textbf{Local tool-result compaction.}
First, we reduce the local verbosity of tool results. Agent trajectories contain many tool-call steps whose outputs are structured or semi-structured, such as JSON payloads, tables, logs, stack traces, command-line outputs, and search results. Instead of treating these outputs as plain text, LiteTrajEval can reuse existing rule-based or format-aware compaction methods to produce shorter but still readable representations. For example, JSON-like outputs can be rendered in token-efficient formats such as TOON~\cite{matveev2026toon}.

\textbf{Cross-step redundancy removal.}
Second, the serializer eliminates redundant information across steps. Scanning the trajectory chronologically, it compares each compacted output with prior steps. If an output is identical or contains near-exact repeated text from earlier steps (e.g., copied code, repeated data records), the redundant segment is replaced with a compact reference marker. Because the step skeleton (input, type, status) remains intact, this de-duplication preserves the execution graph while removing duplicated evidence.

\textbf{Tier-based allocation and intra-step selection.}
After redundancy removal, each remaining step $i$ has a compacted output length $\ell_i$. LiteTrajEval derives a minimum preservation floor based on the step's inferred status tier ($\tau_i \in \{0: \texttt{ok}, 1: \texttt{warning}, 2: \texttt{error}\}$):

\begin{equation}
\label{eq:preservation_floor}
m_i = \min(\ell_i, M_{\tau_i}), \qquad M_0 < M_1 < M_2.
\end{equation}

Here, $M_{\tau_i}$ denotes the configured minimum preservation budgets. The serializer then allocates a final output budget $c_i$ for each step, subject to the constraints:

\begin{equation}
\label{eq:budget_allocation}
m_i \leq c_i \leq \ell_i, \qquad \sum_i c_i \leq B_{\text{out}},
\end{equation}

where $B_{\text{out}}$ is the remaining global budget after reserving space for the task and final answer. The allocation is solved by the waterfall reduction procedure in Algorithm~\ref{alg:budget-serializer}. 
After initializing each step with its full compacted length and tier-dependent preservation floor, the algorithm processes tiers from lower to higher priority and repeatedly shrinks the longest remaining outputs in the current tier toward the next length level or the floor (Lines~3--10). A final hard truncation is used only if the global budget remains violated.

For the assigned budget $c_i$, intra-step shortening is formulated as an evidence selection problem. Using the Maximal Marginal Relevance (MMR) principle~\cite{carbonell1998mmr}, LiteTrajEval constructs a selection anchor $a_i$ from the task and step inputs, and iteratively selects evidence units $u$ to maximize relevance while penalizing redundancy against already selected units $S_i$:

\begin{equation}
\label{eq:mmr_selection}
u^* = \operatorname*{arg\,max}_{u \in U_i \setminus S_i} \left[ \lambda \cdot \mathrm{sim}(u, a_i) - (1-\lambda) \cdot \max_{v \in S_i} \mathrm{sim}(u, v) \right]
\end{equation}

\begin{algorithm}[t]
\caption{Tier-Based Waterfall Budget Allocation}
\label{alg:budget-serializer}
\small
\begin{algorithmic}[1]
\REQUIRE Normalized steps $S$, global budget $B$
\ENSURE Serialized trajectory $T$
\STATE Build $r_i=(h_i,\tilde{o}_i,\tau_i,a_i)$; set $\ell_i=|\tilde{o}_i|$, $c_i=\ell_i$, and $m_i=\min(\ell_i,M_{\tau_i})$ for all $i$.
\STATE Let $\mathcal{R}(\mathbf{c})=\mathrm{Render}(\{h_i,\mathrm{Select}(\tilde{o}_i,c_i,a_i)\}_i)$ and $T\gets\mathcal{R}(\mathbf{c})$.
\FOR{$t\in\{0,1,2\}$}
    \WHILE{$|T|>B$ and $\exists i:\tau_i=t \land c_i>m_i$}
        \STATE $C\gets\{i:\tau_i=t \land c_i>m_i\}$
        \STATE $L\gets\max_{i\in C}c_i$, $G\gets\{i\in C:c_i=L\}$
        \STATE $L'\gets\max(\{c_i:i\in C,c_i<L\}\cup\{0\})$
        \STATE Reduce each $c_i$ for $i\in G$ evenly toward $\max(m_i,L')$.
        \STATE $T\gets\mathcal{R}(\mathbf{c})$
    \ENDWHILE
\ENDFOR
\STATE \textbf{return} $T$ if $|T|\le B$, otherwise $\mathrm{HardTruncate}(T,B)$
\end{algorithmic}
\end{algorithm}

\subsection{Rubric-Guided LLM-as-a-Judge}

The final stage invokes a single rubric-guided LLM judge. Instead of using separate LLM calls for scoring, failure localization, taxonomy classification, and root-cause analysis, LiteTrajEval consolidates these into one structured report: all outputs condition on the same serialized trajectory, so a single call is both sufficient and more cost-efficient.

The judge evaluates both final-output quality and trajectory quality. The rubric covers goal progress and completion, evidence grounding and verification, execution and tool handling, and recovery and adaptive control. The output is a structured JSON report containing rubric scores, candidate failure steps, failure categories, root-cause analysis, and key observations. For a single trace, failure steps and root-cause analysis help developers inspect the relevant execution region. Across many traces, failure categories and rubric scores can be aggregated to identify recurring failure patterns, compare agent versions, and support regression testing.

\section{Evaluation}

We evaluate LiteTrajEval along three practical questions. First, can a lightweight evaluator detect that a failed trajectory contains a problem? Second, once a failure is detected, does the predicted failure region align with human annotations, and how does this alignment compare to a heavier diagnostic baseline? Third, can LiteTrajEval keep token usage, runtime, and evaluation cost predictable under a fixed trajectory budget?

\begin{table*}[t]
\caption{Failure detection rate and group-level alignment on detected samples.}
\centering
\scriptsize
\resizebox{\textwidth}{!}{
\renewcommand{\arraystretch}{1.1}
\begin{tabular}{|l|rrr|rrr|}
\hline
\multirow{2}{*}{\textbf{Solution (Model)}} &
\multicolumn{3}{c|}{\textbf{Magentic-One} (44 samples, 78 failure groups)} &
\multicolumn{3}{c|}{\textbf{$\tau$-retail} (29 samples, 31 failure groups)} \\
\cline{2-7}
& \textbf{Detection Rate}
& \textbf{Align.$\mid$Det.@1}
& \textbf{Align.$\mid$Det.@3}
& \textbf{Detection Rate}
& \textbf{Align.$\mid$Det.@1}
& \textbf{Align.$\mid$Det.@3} \\ 
\hline

AgentRx (GPT-5-mini)
& 97.7 (43/44) & 40.2 (21/77) & 45.4 (35/77)
& 86.2 (25/29) & 51.8 (14/27) & 59.2 (16/27) \\
\hline

\textbf{LiteTrajEval} (No LLM, Status Only)
& 56.8 (25/44) & 45.6 (21/46) & 47.8 (22/46)
& 31.0 (9/29) & 36.3 (4/11) & 54.5 (6/11) \\
\hline

\textbf{LiteTrajEval} (DeepSeek-v3.2)
& 95.4 (42/44) & 56.6 (43/76) & 67.1 (51/76)
& 72.4 (21/29) & 73.9 (17/23) & 73.9 (17/23) \\
\hline

\textbf{LiteTrajEval} (DeepSeek-v4-flash)
& \textbf{100 (44/44)} &  65.4 (51/78) & 73.1 (57/78)
& 82.7 (24/29) & 62.9 (17/27) & 77.8 (21/27) \\
\hline

\textbf{LiteTrajEval} (GPT-5-mini)
& 97.7 (43/44) & 66.2 (51/77) & \textbf{83.1 (64/77)}
& \textbf{89.6 (26/29)} & \textbf{78.6 (22/28)} & \textbf{82.1 (23/28)} \\
\hline

\textbf{LiteTrajEval} (Gemini-3-flash)
& 95.4 (42/44) & \textbf{71.1 (54/76)} & 77.6 (59/76)
& 86.2 (25/29) & 77.8 (21/27) & 77.8 (21/27) \\
\hline
\end{tabular}
}
\label{tab-detection-alignment}
\end{table*}

\begin{table*}[t]
\caption{Trace length, per-sample evaluator token usage, total evaluation cost, and runtime.}
\centering
\begin{threeparttable}
\resizebox{\textwidth}{!}{
\renewcommand{\arraystretch}{1.1}
\begin{tabular}{|l|r|r|r|r|r|r|r|}
\hline
\multirow{2}{*}{\textbf{Metric / Model}}
& \multicolumn{3}{c|}{\textbf{Magentic-One}}
& \multicolumn{3}{c|}{\textbf{$\tau$-retail}}
& \multirow{2}{*}{} \\ \cline{2-7}

& \textbf{Min} & \textbf{Max} & \textbf{Avg}
& \textbf{Min} & \textbf{Max} & \textbf{Avg}
&  \\ \hline

Trace char length
& 9,185 & 360,899 & 82,607
& 20,009 & 76,475 & 35,870
& -- \\ \hline\hline

\multicolumn{7}{|c|}{\textbf{Per-sample evaluator token usage (input / output tokens)}} &  \textbf{Cost (USD)}\tnote{*} \\ \hline

AgentRx (GPT-5-mini)
& 14,327 / 4,737 & 202,650 / 16,415 & 53,995 / 10,851
& 20,036 / 4,457 & 91,889 / 17,297 & 31,882 / 11,773
& \$2.4628 \\ \hline

\textbf{LiteTrajEval} (DeepSeek-v3.2)
& 3,767 / 208 & 15,662 / 919 & 11,130 / 355
& 5,063 / 180 & 11,858 / 430 & 7,202 / 289
& \$0.1681 \\ \hline

\textbf{LiteTrajEval} (DeepSeek-v4-flash)
& 3,765 / 1,195 & 17,052 / 3,715 & 11,728 / 2,321
& 5,061 / 1,288 & 11,856 / 4,096 & 7,200 / 2,725
& \$0.1087 \\ \hline

\textbf{LiteTrajEval} (GPT-5-mini)
& 3,677 / 670 & 16,331 / 1,357 & 11,395 / 1,008
& 4,950 / 1,213 & 11,586 / 3,965 & 7,011 / 2,597
& \$0.4155 \\ \hline

\textbf{LiteTrajEval} (Gemini-3-flash)
& 3,936 / 1,071 & 18,577 / 9,996 & 12,676 / 3,976
& 5,587 / 1,655 & 15,103 / 9,996 & 8,419 / 4,699
& \$1.3346 \\ \hline\hline

\multicolumn{7}{|c|}{\textbf{Per-sample evaluation time (s)}} & \textbf{Total Time (s)} \\ \hline

AgentRx (GPT-5-mini)
& 48.20 & 326.00 & 147.10
& 129.90 & 914.70 & 328.20
& 15,990 \\ \hline

\textbf{LiteTrajEval} (GPT-5-mini)
& 13.00 & 63.00 & 24.34
& 12.00 & 63.00 & 24.34
& 1,840 \\ \hline

\end{tabular}
}
\begin{tablenotes}
\footnotesize
\item[*] \textit{Estimated total cost for evaluating all samples for the corresponding method/model, based on official API prices as of June 2026.}
\end{tablenotes} 
\end{threeparttable}
\label{tab-token}
\end{table*}

\subsection{Datasets, Baseline, and Metrics}
We evaluate on two public failed-trajectory settings derived from the AgentRx benchmark: Magentic-One-style trajectories (44 samples, 78 normalized failure groups) and $\tau$-retail trajectories (29 samples, 31 normalized failure groups), the latter involving policy-constrained interactions among an agent, a user, and external tools.

We compare against AgentRx~\cite{barke2026agentrx}, the closest baseline because it also targets failure localization and provides human failure annotations. We run AgentRx in a cost-minimized configuration; even so, it requires multiple LLM calls per sample, while LiteTrajEval uses one judge call after preprocessing and budget-aware serialization. We also include a \textit{No LLM, Status Only} variant using only rule-based step status signals.

We report two metrics. \textit{Detection Rate} is the fraction of failed samples for which the evaluator reports at least one candidate failure step, i.e., a non-empty \texttt{failure\_steps} field for LiteTrajEval. Since all samples are failed trajectories, this measures failure recall.

\textit{Align.$\mid$Det.@k} measures group-level agreement with human annotations among detected samples. Because one causal failure may span adjacent steps, we normalize human annotations into \textit{failure groups} $G=\{g_1,\ldots,g_m\}$. A predicted step $p$ matches $G$ at tolerance $k$ if $\min_{g \in G}|p-g| \leq k$; we report $k=1$ and $k=3$:

\begin{equation}
\label{eq:alignment_metric}
\text{Align.$\mid$Det.@k} = \frac{N_{\text{matched failure groups at tolerance } k}}{N_{\text{failure groups in detected samples}}}
\end{equation}

We use \textit{alignment} rather than \textit{accuracy} because human annotations are reference regions rather than unique ground truth; predictions outside the tolerance window may still identify useful related symptoms or consequences. We also report evaluator token usage, estimated cost, and runtime: for AgentRx these aggregate all pipeline LLM calls, while for LiteTrajEval they correspond to the single judge call under a fixed 50,000-character trajectory budget.

\begin{figure*}[!t]
\centering
\includegraphics[width=\linewidth]{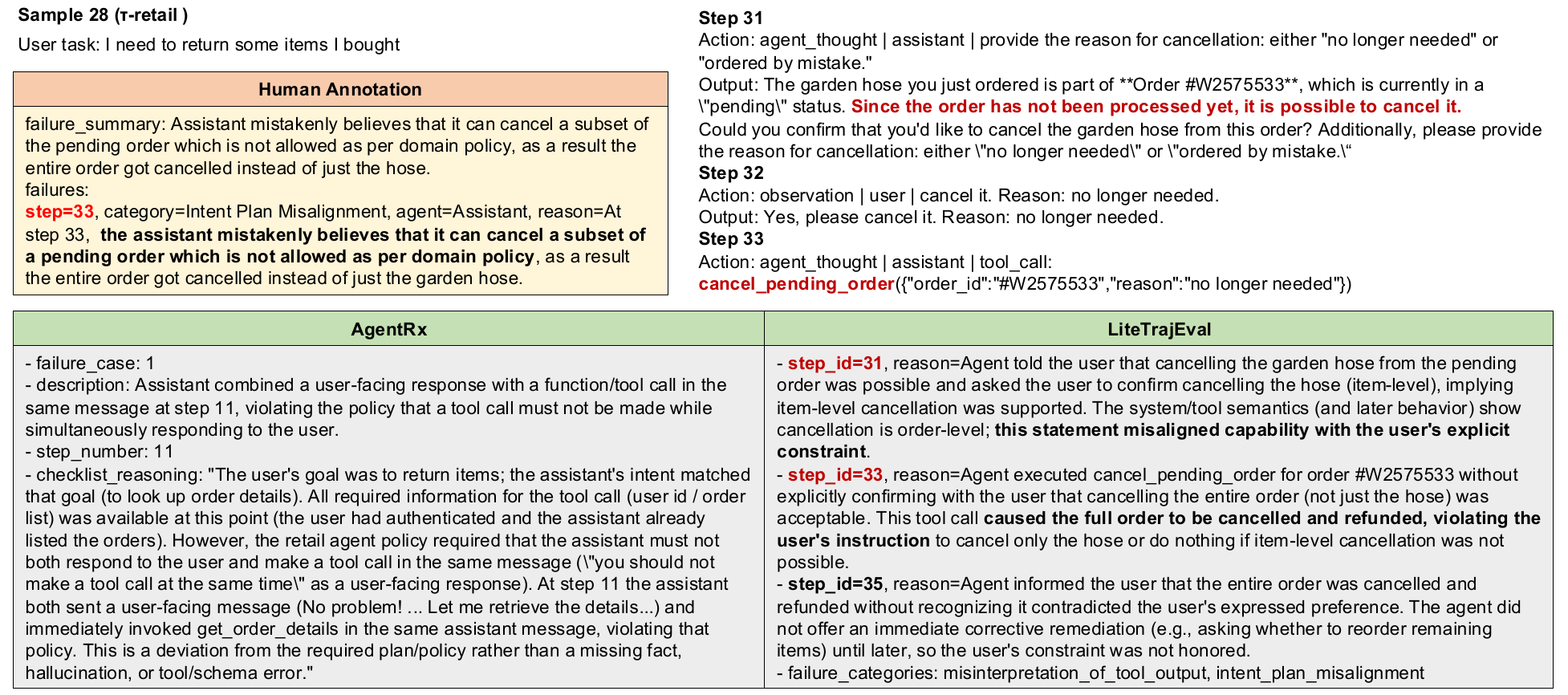}
\caption{Qualitative comparison on a $\tau$-retail failure case. AgentRx localizes an unrelated policy violation, mismatching the human annotation, whereas LiteTrajEval identifies the order-level cancellation error and its downstream effects.}
\label{fig:case-study}
\end{figure*}

\subsection{RQ1: Can Lightweight Judges Detect and Localize Failed Trajectories?}
Table~\ref{tab-detection-alignment} shows that rule-based status signals alone, which reflect the maximum detection capability of our offline rule generation stage, are not sufficient for reliable trajectory evaluation. Figure~\ref{fig:case-study} illustrates such a case: the failure is not exposed by a local error status, but requires reasoning over the user's constraint, tool semantics, and downstream cancellation outcome. This is consistent with agent execution: local error messages do not always indicate root causes, since agents may recover through retries or alternative paths, and early reasoning mistakes may only become visible later as downstream failures. The \textit{No LLM, Status Only} variant detects substantially fewer failed trajectories than any LLM-based configuration (56.8\% on Magentic-One and 31.0\% on $\tau$-retail), so many failed trajectories would not be surfaced for diagnosis. While its alignment on detected samples is not negligible, especially at @1, its low detection coverage is the primary limitation.

LLM-based LiteTrajEval substantially improves detection coverage across both datasets while also achieving stronger group-level alignment on detected samples than the rule-based variant. This gap suggests that the judge uses information beyond rule-inferred step statuses alone, including model reasoning, recovery behavior, failure propagation, and whether a later visible error is a symptom of an earlier mistake. Compared with AgentRx, the detection picture is mixed across judge models. GPT-5-mini, Gemini-3-flash, and DeepSeek-v4-flash achieve detection rates comparable to or higher than AgentRx on both datasets. DeepSeek-v3.2, evaluated without reasoning mode, shows a lower detection rate on $\tau$-retail (72.4\% versus AgentRx's 86.2\%), indicating that detection coverage on policy-constrained, interaction-heavy trajectories benefits from reasoning-capable judges.

Despite this mixed detection picture, LiteTrajEval consistently and substantially improves conditional alignment with human annotations once a failure is detected. Across all reported judge models, Alignment$\mid$Detected@3 exceeds AgentRx by roughly 20--35 percentage points on both datasets. This indicates that, when a failed trajectory is surfaced, LiteTrajEval more often points to a failure region that agrees with the human reference annotations than AgentRx does, while preserving a lightweight, single-call online evaluation path.

The lower detection coverage on $\tau$-retail also reflects the nature of some failures in this dataset. Several undetected cases involve user-side behavior, such as incorrect user information or missing confirmation, rather than an obvious tool-execution or agent-reasoning error. Since the current judge prompt focuses mainly on tool execution, model reasoning, and agent reactions, these interaction-state failures are harder to identify and represent a natural direction for future prompt refinement.

\subsection{RQ2: Does LiteTrajEval Keep Evaluation Lightweight?}

Table~\ref{tab-token} reports raw trace lengths, per-sample evaluator token usage, and total evaluation cost. For AgentRx, token usage is first aggregated across all LLM calls required for a sample, and the reported min, max, and average are then computed over samples. For LiteTrajEval, each sample requires one online judge call, so the reported token usage directly corresponds to that single call. The cost column reports the total estimated cost of running the full evaluation set, including all 44 Magentic-One samples and 29 $\tau$-retail samples.

The cost comparison highlights the effect of the lightweight design. Under the same GPT-5-mini judge model, LiteTrajEval reduces total evaluation cost from \$2.4628 to \$0.4155, a reduction of about 6$\times$. When using DeepSeek-v4-flash, total cost decreases further to \$0.1087, corresponding to about 22.7$\times$ lower cost than AgentRx with GPT-5-mini. Importantly, this cost reduction does not come at the expense of localization quality: DeepSeek-v4-flash maintains strong group-level alignment on detected samples across both datasets (73.1\% and 77.8\% at @3), substantially above AgentRx (45.4\% and 59.2\%), despite a detection rate on $\tau$-retail that is slightly below AgentRx's. GPT-5-mini provides the strongest overall combination of detection and alignment on $\tau$-retail among all evaluated configurations.

LiteTrajEval also runs faster and more consistently: total evaluation time drops from 15,990s to 1,840s (8.7$\times$), with per-sample runtime confined to 12--63 seconds versus AgentRx's 48--914 seconds, as shown in Table~\ref{tab-token}. This stability follows from the fixed-budget, single-call design.

\section{Lessons from Production Deployment}

LiteTrajEval has been integrated into the evaluation harness of our enterprise agentic platform. Our first pilot targets an internal SRE agent that performs automated root-cause analysis (RCA) for incidents on a large-scale cloud platform, executing hundreds of ReAct-style trajectories per day, each with several dozen steps. The lightweight design keeps evaluation at this volume practical: the full daily traffic can be evaluated, and re-evaluated after prompt or configuration changes, within routine cost and latency budgets. Tool outputs are highly skewed, with a small number of diagnostic calls (e.g., log queries) returning very large payloads while other steps are short. In practice, compacting the largest-output steps accounts for most of the reduction in serialized trace length, confirming the design of our serialization strategy. This compaction remains structural rather than semantic; hence, applying LLM-based summarization selectively to these outlier steps without adding a per-step LLM call is a natural extension.

For judging, small open-weight, self-hosted models (e.g., DeepSeek-v4-flash, Qwen3.5-35B) were sufficient to stay within production latency and cost budgets; no frontier model was required. In preliminary manual inspection, the surfaced failure steps were plausible and remained largely stable across judge models. Since production traffic carries no step-level ground-truth labels, we report this as a qualitative observation; a lightweight human-in-the-loop labeling process and calibration assessment across judge models remain ongoing work.

Finally, the judge is loosely coupled with preprocessing and serialization. Our ongoing experiments in skill auto-optimization and agent memory reuse the same pipeline, requiring only a different rubric and output schema. This suggests that lightweight trajectory evaluation can serve as a reusable building block for broader agent-improvement workflows, not only failure diagnosis.

\section{Conclusion}
We presented LiteTrajEval, a lightweight architecture for budget-bounded trajectory evaluation of production LLM agents. LiteTrajEval combines offline rule-profile generation, deterministic trajectory preprocessing, fixed-budget serialization, and a single rubric-guided LLM judge call. On public Magentic-One-style and $\tau$-retail trajectories, LiteTrajEval improves failure-localization alignment with human annotations while reducing evaluation cost and runtime compared with a strong multi-call diagnosis baseline. By preserving the global execution skeleton and removing redundant low-value content, LiteTrajEval enables repeated agent debugging, regression testing, and offline failure analysis within a predictable evaluation budget.







\begin{thebibliography}{00}

\bibitem{yao2024taubench}
S. Yao, N. Shinn, P. Razavi, and K. R. Narasimhan,
``$\tau$-bench: A Benchmark for Tool-Agent-User Interaction in Real-World Domains,''
in \textit{Proc. International Conference on Learning Representations (ICLR)}, 2025.

\bibitem{chang2024agentboard}
C. Ma et~al.,
``AgentBoard: An Analytical Evaluation Board of Multi-turn LLM Agents,''
in \textit{Advances in Neural Information Processing Systems 37 (NeurIPS 2024), Datasets and Benchmarks Track}, 2024.

\bibitem{chen2024teval}
Z. Chen et~al.,
``T-Eval: Evaluating the Tool Utilization Capability of Large Language Models Step by Step,''
in \textit{Proc. 62nd Annual Meeting of the Association for Computational Linguistics (ACL), Volume 1: Long Papers}, Bangkok, Thailand, 2024, pp. 9510--9529.

\bibitem{holistic2026}
N. Madvil et~al.,
``Holistic Evaluation and Failure Diagnosis of AI Agents,''
\textit{arXiv preprint arXiv:2605.14865}, 2026.

\bibitem{agenteval2026}
D. Guo, J. Wu, and S. M. Yiu,
``AgentEval: DAG-Structured Step-Level Evaluation for Agentic Workflows with Error Propagation Tracking,''
\textit{arXiv preprint arXiv:2604.23581}, 2026.



\bibitem{kim2025trace}
W. Kim et~al.,
``Beyond the Final Answer: Evaluating the Reasoning Trajectories of Tool-Augmented Agents,''
\textit{arXiv preprint:2510.02837}.

\bibitem{lu2025agentrewardbench}
X. H. L\`u et~al., ``AgentRewardBench: Evaluating Automatic Evaluations of Web Agent Trajectories,''
\textit{arXiv preprint arXiv:2504.08942}.

\bibitem{zhuge2025agentjudge}
M. Zhuge et~al., ``Agent-as-a-Judge: Evaluate Agents with Agents,''
in \textit{Proc. 42nd International Conference on Machine Learning (ICML)},
PMLR, vol. 267, 2025, pp. 80569--80611.

\bibitem{ou2025agentdiagnose}
T. Ou, W. Guo, A. Gandhi, G. Neubig, and X. Yue,
``AgentDiagnose: An Open Toolkit for Diagnosing LLM Agent Trajectories,''
in \textit{Proc. 2025 Conference on Empirical Methods in Natural Language Processing: System Demonstrations},
Suzhou, China, 2025, pp. 207--215.

\bibitem{liu2023lost}
N. F. Liu et~al.,
``Lost in the Middle: How Language Models Use Long Contexts,''
\textit{arXiv preprint:2307.03172}, 2023.


\bibitem{barke2026agentrx}
S. Barke, A. Goyal, A. Khare, A. Singh, S. Nath, and C. Bansal,
``AgentRx: Diagnosing AI Agent Failures from Execution Trajectories,''
\textit{arXiv preprint:2602.02475}, 2026.

\bibitem{matveev2026toon}
I. Matveev,
``Token-Oriented Object Notation vs JSON: A Benchmark of Plain and Constrained Decoding Generation,''
\textit{arXiv preprint:2603.03306}, 2026.

\bibitem{carbonell1998mmr}
J. Carbonell and J. Goldstein,
``The Use of MMR, Diversity-Based Reranking for Reordering Documents and Producing Summaries,''
in \textit{Proceedings of the 21st Annual International ACM SIGIR Conference on Research and Development in Information Retrieval}, 1998.

\end{thebibliography}
\end{document}